\documentclass[letterpaper, 10 pt, journal, twoside]{ieeeconf}
\IEEEoverridecommandlockouts
\usepackage{graphicx}
\usepackage{amsmath}
\usepackage{amssymb}
\usepackage{makecell}
\usepackage[activate={true,nocompatibility},final,tracking=true,kerning=true,spacing=true]{microtype}
\usepackage{tabularx}
\usepackage{doi}
\usepackage{cite}
\hypersetup{hidelinks}
\usepackage{stfloats}
\usepackage[table, x11names]{xcolor}
\usepackage{array, boldline} 

\usepackage{nicematrix}
\usepackage{adjustbox}
\usepackage{etoolbox}

\makeatletter
\renewcommand{\section}{%
  \@startsection{section}{1}{\z@}%
  {1.5ex plus 0.5ex minus 0.5ex}
  {0.5ex}
  {\normalfont\normalsize\centering\scshape}%
}

\renewcommand{\subsection}{%
  \@startsection{subsection}{2}{\z@}%
  {1.0ex plus 0.5ex minus 0.3ex}
  {0.4ex}
  {\normalfont\normalsize\itshape}%
}

\patchcmd{\@makecaption}
  {\begin{center}{\footnotesize #1}\\{\footnotesize\scshape #2}\end{center}}
  {\begin{center}{\footnotesize #1: \scshape #2}\end{center}}
  {}{\typeout{Table caption patch failed}}
\makeatother

\title{\LARGE Plug, Play and Comply: Plugin-Based Compliant Control in ROS 2}

\author{Mihael Simoni\v{c}$^{1}$ and Xiaocong Li$^{1,2,*}$%
\thanks{$^{1}$ College of Information Science and Technology, Eastern Institute of Technology, Ningbo, Ningbo 315200, China}%
\thanks{$^{2}$ Zhejiang Key Laboratory of Industrial Intelligence and Digital Twin,
Eastern Institute of Technology, Ningbo, Ningbo 315200, China}%
\thanks{Email: msimonic@eitech.edu.cn, xiaocongli@eitech.edu.cn. Corresponding author: Xiaocong Li.}%
}
\begin{document}

\maketitle

\begin{abstract}
The paper presents a robot-agnostic framework for implementing and deploying compliant-control algorithms across different manipulators. It extends the ROS control ecosystem with a plugin-based execution architecture that separates ROS controller boilerplate from control-law implementation. 
Standardized joint and Cartesian command interfaces provide a consistent message-level interface to higher-level applications, supporting motion references, stiffness and damping gains, nullspace targets, desired interaction wrenches, and independent feedforward inputs. Robot kinematics and dynamics are computed from URDF models using Pinocchio. The common plugin interface also supports control implementations generated through block-diagram-based pipelines. A real-robot curved-surface following experiment demonstrates the utility of online task-aligned compliance, while simulations demonstrate portability across manipulators with different kinematic and dynamic characteristics. The complete framework, including reference controllers, high-level task interfaces, and example configurations, is available as open-source software.
\end{abstract}


\section{Introduction}

Robot manipulation systems increasingly integrate high-level task generation with low-level execution. Vision-language-action (VLA) models broaden the range of commands that must be translated into physical robot behavior \cite{Shao2025VLMSurvey}, while teleoperation, residual human correction, and shared autonomy require predictable responses to uncertain human input and environmental contact \cite{Xu2025CRDagger}. These systems may produce low-frequency action chunks, task-level corrections, or semantic motion objectives that differ substantially from the inputs expected by the low-level control loop. The low-level controller must translate these commands into smooth motion while regulating physical interaction.

Contact-rich tasks such as peg insertion, connector mating, and surface following commonly rely on impedance or admittance control \cite{Tsuji2026ContactRichSurvey} and require compliance to adapt during execution \cite{Buchli2011LearningVIC,Kronander2016LearningCompliantManipulation}. Search and alignment may call for low stiffness, whereas subsequent contact phases may require greater stiffness in selected directions and sufficient damping to limit impacts and oscillations. Recent work also indicates that controller gains influence imitation learning, reinforcement learning, and sim-to-real transfer \cite{bronars2026tunelearncontrollergains}. VLA-based systems have begun generating context-dependent stiffness and damping alongside motion commands \cite{zhang2026compliantvla}. Together, these requirements motivate command interfaces that allow high-level systems to update motion and compliance parameters online.

The ROS control framework provides robot-agnostic access to hardware and simulation through common state and command interfaces \cite{Chitta2017RosControl,ros2control}. Its standard controllers primarily support position, velocity, effort, and trajectory execution. Compliant manipulation requires richer command formulation that combines motion references, stiffness and damping, feedforward terms, nullspace targets, and task-frame information. Existing compliant-control packages provide subsets of this functionality, but differ in command semantics and robot-model access; some also depend on particular robots, dynamics libraries, or vendor APIs. These differences complicate reuse of control laws and high-level command sources across platforms.

To address this integration gap, we propose a wrapper--plugin architecture using standardized command interfaces. The contributions are: (1) a robot-agnostic architecture separating ROS controller deployment from interchangeable, ROS-independent control laws; (2) joint- and Cartesian-space command interfaces supporting concurrent online updates of motion references and compliance parameters, including full translational and rotational stiffness and damping matrices for task-aligned compliance; and (3) reference joint- and Cartesian-space impedance controllers, together with tooling for both conventional C++ and block-scheme-based plugin development.

The remainder of the paper reviews related ROS compliant-control packages, presents the architecture and command interfaces, and describes the reference controllers and Simulink-to-plugin workflow. Evaluation covers deployment across multiple robot models and a hardware curved-fixture-following experiment with task-aligned compliance.

\section{Related Work}

\newcommand{\repo}[3]{\makecell[l]{\href{#1}{\scriptsize{#2}}\\[-0.3mm]\href{#1}{\scriptsize{#3}}}}
\newcommand{\repoFZI}{\repo{https://github.com/fzi-forschungszentrum-informatik/cartesian_controllers}{fzi-forschungszentrum-informatik/}{cartesian\_controllers}}
\newcommand{\repoICube}{\repo{https://github.com/ICube-Robotics/cartesian_controllers_ros2}{ICube-Robotics/}{cartesian\_controllers\_ros2}}
\newcommand{\repoMayr}{\repo{https://github.com/matthias-mayr/Cartesian-Impedance-Controller}{matthias-mayr/Cartesian-}{Impedance-Controller}}
\newcommand{\repoROSImp}{\repo{https://github.com/qleonardolp/ros2_impedance_controller}{qleonardolp/}{ros2\_impedance\_controller}}
\newcommand{\repoDeplazes}{\repo{https://github.com/CurdinDeplazes/cartesian_impedance_control}{CurdinDeplazes/}{cartesian\_impedance\_control}}
\newcommand{\repoSERL}{\repo{https://github.com/rail-berkeley/serl_franka_controllers}{rail-berkeley/}{serl\_franka\_controllers}}
\newcommand{\repoKUKA}{\repo{https://github.com/idra-lab/kuka_lbr_control}{idra-lab/}{kuka\_lbr\_control}}
\newcommand{\repoIJS}{\repo{https://github.com/abr-ijs/ijs_controllers}{abr-ijs/}{ijs\_controllers}}
\newcommand{\repoCRISP}{\repo{https://github.com/utiasDSL/crisp_controllers}{utiasDSL/}{crisp\_controllers}}
\newcommand{\repoAALCC}{\repo{https://github.com/applied-ai-lab/compliant_controllers}{applied-ai-lab/}{compliant\_controllers}}
\newcommand{\ours}{\hlineB{0}\rowcolor{SeaGreen3!10!} Compliant Cntrl. (OUR)}

\newcommand{\initupdate}{\fbox{\scriptsize I}}
\newcommand{\paramupdate}{\fbox{\scriptsize P}}
\newcommand{\topicupdate}{\fbox{\scriptsize T}}
\newcommand{\serviceupdate}{\fbox{\scriptsize S}}
\newcommand{\combined}{\fbox{\scriptsize C}}

\newcommand{\rcell}[1]{\rotatebox[origin=c]{90}{\scriptsize\makecell[l]{#1}}}
\newcommand{\ccell}[1]{\makecell[c]{#1}}
\newcommand{\lcell}[1]{\makecell[l]{#1}}

\begin{table*}[t]
\vspace{-1em}
\centering
\scriptsize
\caption{Feature comparison of representative compliant controllers in ROS 2.}
\label{tab:controllers_comparison}
\setlength{\tabcolsep}{2pt}
\begin{NiceTabular*}{\textwidth}{@{} l @{\extracolsep{\fill}} cccccccccccc @{}}[color-inside]
\hline 
\lcell{Controller package / repository} &
\ccell{Dynamics\\library\\interface} & 
\ccell{Robot\\agnostic} & 
\ccell{Simulation\\support} & 
\ccell{Joint\\limits\\handling} & 
\ccell{Nullspace\\control} & 
\ccell{Feed-\\forward\\torque/wrench} & 
\ccell{Friction\\comp.} & 
\ccell{Exchange\\control\\plugin} & 
\ccell{Combined\\ref+gains\\update} & 
\ccell{Config.\\stiffness\\frame} & 
\ccell{Full\\rotation} & 
\ccell{Online\\gain\\update} \\
\hline
\repoFZI      & custom      & \checkmark & \checkmark & \checkmark & --         & \checkmark & --         & --         & --         & \checkmark & \checkmark & P          \\
\repoICube    & KDL         & \checkmark & --         & \checkmark & \checkmark & \checkmark & --         & \ccell{ROS-\\dependent} & \checkmark & \checkmark & \checkmark & P, C       \\
\repoMayr     & RBDyn       & \checkmark & \checkmark & \checkmark & \checkmark & \checkmark & --         & --         & --         & --         & --         & P, T       \\
\repoROSImp   & Pinocchio   & \checkmark & \checkmark & --         & --         & --         & --         & --         & --         & --         & --         & -- (Init)       \\
\repoDeplazes & libfranka   & --         & \checkmark & --         & --         & \checkmark & --         & --         & --         & --         & --         & -- (Mode)     \\
\repoKUKA     & KUKA        & --         & --         & --         & --         & \checkmark & --         & --         & --         & --         & --         & -- (Init)       \\
\repoCRISP    & Pinocchio   & \checkmark & \checkmark & \checkmark & \checkmark & \checkmark & \checkmark & --         & --         & --         & --         & P, T       \\
\ours         & Pinocchio   & \checkmark & \checkmark & \checkmark & \checkmark & \checkmark & \checkmark & \ccell{ROS-\\indep.} & \checkmark & \checkmark & \checkmark & C, T       \\
\hline
\end{NiceTabular*}
\begin{minipage}{\textwidth}
\vspace{2px}
\raggedright\scriptsize
\textit{Legend:} \checkmark denotes support and -- denotes no support. \textit{Online gain update paths:} P/T - using separate parameter/topic interface; C - combined motion/gain command topic; Mode - switch between preconfigured gains; Init - startup configuration only. \textit{Exchangeable control plugin:} ROS-dependent / ROS-independent indicate whether the plugin interface depends on ROS types. \textit{Frame / rotation:} The two columns indicate a configurable compliance frame and support for full, non-diagonal stiffness rotation, respectively. 
\vspace{-1em}
\end{minipage}
\end{table*}

ROS control supports controller reuse through common hardware interfaces, URDF descriptions, and a shared controller manager. Compliant-control packages extend this foundation with motion and compliance commands. Table~\ref{tab:controllers_comparison} compares relavant joint- and Cartesian-space implementations in terms of portability, control features, and extensibility (as of October 2026).
Several packages use vendor-provided dynamics and hardware integration but are not readily transferable across platforms. Robot-agnostic frameworks improve portability, yet differ in model dependencies, simulation support, nullspace handling, feedforward commands, and extensibility. 
CRISP provides a ROS~2 framework supporting simulation, nullspace regulation, feedforward commands, and friction compensation, with gain updates exposed through a separate interface \cite{SanJosePro2025CRISP}. ICube supports loadable effort-based impedance and position/velocity-based admittance rules, with stiffness, damping, and inertia specified at compliant-frame trajectory points \cite{icube_repo}. Its rule interface includes ROS controller, message, timing, and parameter types; our control-law interface is ROS-independent.
Our matrix-based interface also lets a task-level source compute and stream full stiffness and damping matrices from contact geometry, so compliance directions can follow a changing surface without requiring a corresponding robot frame in TF.
Our framework combines this interface boundary with standardized joint and Cartesian commands and URDF-backed model access to support reuse across robots.

\section{Proposed Framework}

\begin{figure*}[b]
    \centering
    \includegraphics[width=\linewidth]{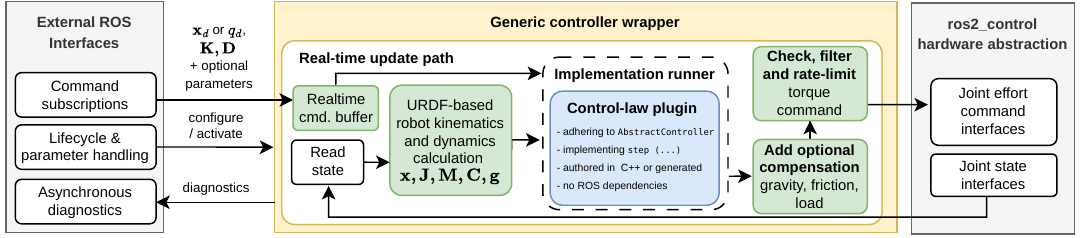}
    \vspace*{-5mm}
    \caption{Architecture and command execution flow of the proposed framework. The wrapper loads configuration parameters, initializes the model and compensation modules, and loads the selected implementation library. During each update, the wrapper reads joint states, updates the URDF-backed robot model, retrieves the latest buffered Cartesian or joint command, invokes the selected control-law plugin, adds configured compensation, applies filtering, finite-value checks, and torque-rate limits, and writes the resulting effort command. Plugins are interchangeable without wrapper recompilation.}
    \label{fig:framework_architecture}
\end{figure*}

The framework consists of a ROS controller wrapper, a ROS-independent control-law plugin interface, a URDF-backed robot-model library, and a command-message package. Joint and Cartesian wrappers share the same execution structure while accepting different reference commands and requesting different model quantities. The wrappers handle hardware interaction, command buffering, compensation, and torque validation, whereas plugins implement the control laws. The robot-model library uses Pinocchio \cite{carpentier-sii19} to provide kinematic and dynamic quantities. The complete software implementations and executable examples are provided in the supplementary resources. 

Before real-time control begins, each wrapper initializes the robot model and compensation modules from the configured robot, joint, and frame parameters, and dynamically loads the selected control-law plugin. Alternative control laws can be deployed by selecting a different implementation library through reconfiguration. Parameters are exposed through standard ROS parameter services and applied during controller initialization. 
The Cartesian wrapper selects the controlled frame from parameters, defaulting to the robot flange, when no end-effector is configured. 
When the controller is started, the current robot state is used to initialize the first Cartesian pose or joint target, and control begins only after a valid initial output is obtained.

Each real-time update follows the sequence shown in Fig.~\ref{fig:framework_architecture}: the wrapper reads the robot state, updates the required model quantities, retrieves the latest buffered command, and invokes the selected plugin to compute joint torques. Configured gravity, friction, payload, and soft joint-limit compensation are subsequently applied, followed by filtering and rate limiting. 
Restarting the controller resets the reference to the measured state to avoid discontinuities. Diagnostics are handled asynchronously. 


The command interfaces decouple reference generation from controller execution, allowing integration with external task-generation frameworks such as RBS Python \cite{Simonic2025RobotBlockSetPython}. The Cartesian message carries task-space references, translational and rotational compliance, desired interaction wrench, feedforward terms, and optional nullspace regulation. The joint message provides the corresponding joint-space references, gains, and feedforward torque. 
By exposing these quantities directly rather than prescribing a specific compliance formulation, the interfaces accommodate different control laws and task-dependent parameterizations. In particular, the full Cartesian compliance matrices can be adjusted online, allowing compliance directions to adapt to task geometry or environmental constraints.

\section{Reference Implementations}
\label{sec:reference_plugins}

The framework includes reference joint- and Cartesian-space impedance-control plugins that demonstrate how different control laws can be implemented behind the same ROS-independent plugin interface. 
Both compute joint torques from the commands and robot states supplied by their wrappers.
The plugin boundary is not limited to the formulations presented bellow, alternative approaches such as \cite{Haddadin2024UFIC} can be accommodated without changing the ROS wrapper.

\subsection{Joint Impedance Controller Plugin}

The joint impedance controller implements a virtual spring--damper behavior \cite{Hogan1985ImpedancePartII}:
\begin{equation}
\boldsymbol{\tau}_{\mathrm{ji}}=
\mathbf{K}_q(\boldsymbol{q}_d-\boldsymbol{q})+
\mathbf{D}_q(\dot{\boldsymbol{q}}_d-\dot{\boldsymbol{q}}) + \boldsymbol{\tau}_{\mathit{ff}}.
\label{eq:ji}
\end{equation}
where $\boldsymbol{q}$ and $\boldsymbol{q}_d$ are the measured and desired joint positions, $\mathbf{K}_q$ and $\mathbf{D}_q$ are joint stiffness and damping gains, and $\boldsymbol{\tau}_\mathit{ff}$ denotes optional feedforward torques.

\subsection{Cartesian Impedance Controller Plugin with Task-Aligned Compliance}
\label{sec:cartesian-impedance-plugin}

The Cartesian impedance plugin computes joint torques from commanded pose, velocity, and compliance parameters. Its formulation supports task-aligned compliance by allowing stiffness and damping directions to adapt to task geometry. The commanded torques are
\begin{equation}
\begin{aligned}
\boldsymbol{\tau}_{\mathrm{ci}}
=\mathbf{J}^T\Big[
&-\mathbf{K}
\begin{bmatrix}
\mathbf{p}-\mathbf{p}_d\\
-\mathbf{R}\boldsymbol{\phi}
\end{bmatrix}
-\mathbf{D}(\mathbf{J}\dot{\boldsymbol{q}}-\mathbf{v}_d) +\mathbf{w}_{\mathit{ff}}
\Big]+\boldsymbol{\tau}_{\mathit{ns}},
\end{aligned}\label{eq:ci}
\end{equation}
where $(\mathbf{p},\mathbf{R})$ is the selected frame's pose, $\mathbf{J}$ its geometric Jacobian, and $\mathbf{v}_d$ the desired twist, all expressed in the robot base frame. The feedforward wrench $\mathbf{w}_{\mathrm{ff}}$ is optional, and $\boldsymbol{\tau}_{\mathit{ns}}$ denotes the nullspace torque. The orientation error is $\boldsymbol{\phi}=2\log_q(\boldsymbol{\eta}^{-1}\otimes\boldsymbol{\eta}_d)$, where $\boldsymbol{\eta}$ and $\boldsymbol{\eta}_d$ are the current and desired unit quaternions, respectively. For redundant manipulators, the nullspace term follows~\cite{Dietrich2015}. The command interface accepts translational and rotational stiffness and damping blocks:
\begin{equation}
\mathbf{K}=\operatorname{diag}(\mathbf{K}_{\mathrm{pos}},\mathbf{K}_{\mathrm{ori}}),
\qquad
\mathbf{D}=\operatorname{diag}(\mathbf{D}_{\mathrm{pos}},\mathbf{D}_{\mathrm{ori}}).
\end{equation}
Each $3\times3$ block is forwarded in full, allowing arbitrary orientations of the principal compliance directions. This way, diagonal stiffness and damping matrices defined in a task frame can be rotated into the robot base frame and supplied directly through the command interface.

\subsection{Compensation, Joint Limit Handling, and Filtering}

Gravity, friction, payload, and soft joint-limit compensation are handled by the wrapper independently of the selected control-law plugin. Each compensation term can be enabled or disabled depending on the capabilities and built-in compensation of the target robot platform. Friction compensation uses experimentally identified robot dynamics and friction parameters \cite{Gaz2019PandaIdentification}. Soft joint-limit handling uses virtual spring--damper torques that become active within configurable margins of the mechanical joint limits, opposing further motion toward the limits \cite{JointLimitImpedance2017}. The commanded torques are assembled as:
\begin{equation}
\boldsymbol{\tau}_{\mathrm{cmd}} =
\boldsymbol{\tau}_{\mathrm{impl}} +
\boldsymbol{\tau}_{g} +
\boldsymbol{\tau}_{\mathrm{frc}} +
\boldsymbol{\tau}_{\mathrm{load}} +
\boldsymbol{\tau}_{\mathrm{lim}},
\end{equation}
where $\boldsymbol{\tau}_{\mathrm{impl}}$ denotes the plugin output (\ref{eq:ji}) or (\ref{eq:ci}) and the remaining terms represent gravity, friction, payload, and joint-limit compensation, respectively. As soft joint-limit torques act directly in joint space, they may influence task-space tracking near joint limits. Note that plugins implementing model-based control laws, such as operational-space control \cite{khatib1987unified}, can directly access robot dynamics through the model interface rather than relying on wrapper-level compensation.

Before hardware execution, the combined torques are smoothed using a configurable first-order low-pass filter and subjected to element-wise torque-rate limiting to prevent abrupt changes in commanded torque.

\section{Simulink-to-Plugin Pipeline}
\label{sec:simulink_pipeline}

\begin{figure}[b]
    \centering
    \includegraphics[width=0.97\linewidth]{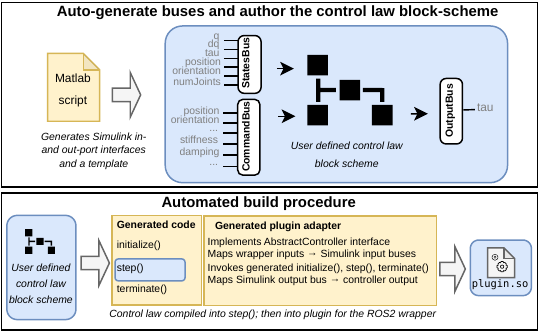}
    \vspace{-3mm}
    \caption{Simulink to plugin workflow. Generated buses connect the control model to the wrapper. The compiled control law conforms to the common plugin interface and can be executed by the  implemetnation runner, cf. Fig.~\ref{fig:framework_architecture}.}
   \label{fig:simulink_pipeline}
\end{figure}

Developing and validating low-level robot controllers directly within a ROS package can require substantial middleware-specific code, creating barriers for block-scheme-based development. 
The wrapper--plugin separation introduced in this work decouples controller design from middleware-specific integration and deployment. As illustrated in Fig.~\ref{fig:simulink_pipeline}, the pipeline converts a structured Simulink control model into a loadable plugin for the proposed wrapper.
The model uses structured command and state buses as inputs and a torque-command bus as output, matching the wrapper's internal data representation while keeping the control model independent of ROS. The pipeline automatically generates the required bus structures and model template, allowing users to implement only the control law before compiling it into a shared library conforming to the wrapper interface.

Unlike robot-specific Simulink toolboxes tied to vendor APIs, the proposed pipeline is robot-agnostic and integrates with the ROS control ecosystem through the wrapper layer. The same generated plugin can execute on different robots or simulators as long as the hardware exposes compatible ROS control interfaces and the configuration provides the appropriate robot description, joints, and frames.
\section{Experimental Evaluation}

The framework was evaluated for real-time feasibility, cross-platform portability, and task-aligned compliance. All experiments used the Cartesian impedance plugin on ROS~2 Humble. Hardware experiments were performed on a Franka FR3 at 1~kHz through the Franka Control Interface (FCI). Task commands were generated using RBS Python.

\subsection{Reference Implementation Timing}

The computational feasibility at a 1 kHz control rate was benchmarked on an Intel Core i7-14700 over 10,000 iterations using the FR3 model. The Cartesian impedance plugin required a mean of 3.82~$\mu$s (max. 21.83~$\mu$s), increasing to 4.49~$\mu$s (max. 22.97~$\mu$s) when including wrapper overhead for model updates and pose extraction. 

\subsection{Cross-Platform Deployment}

\begin{figure}[b]
  \centering
  \begin{minipage}[b]{0.24\linewidth}
    \centering
    \includegraphics[width=\linewidth]{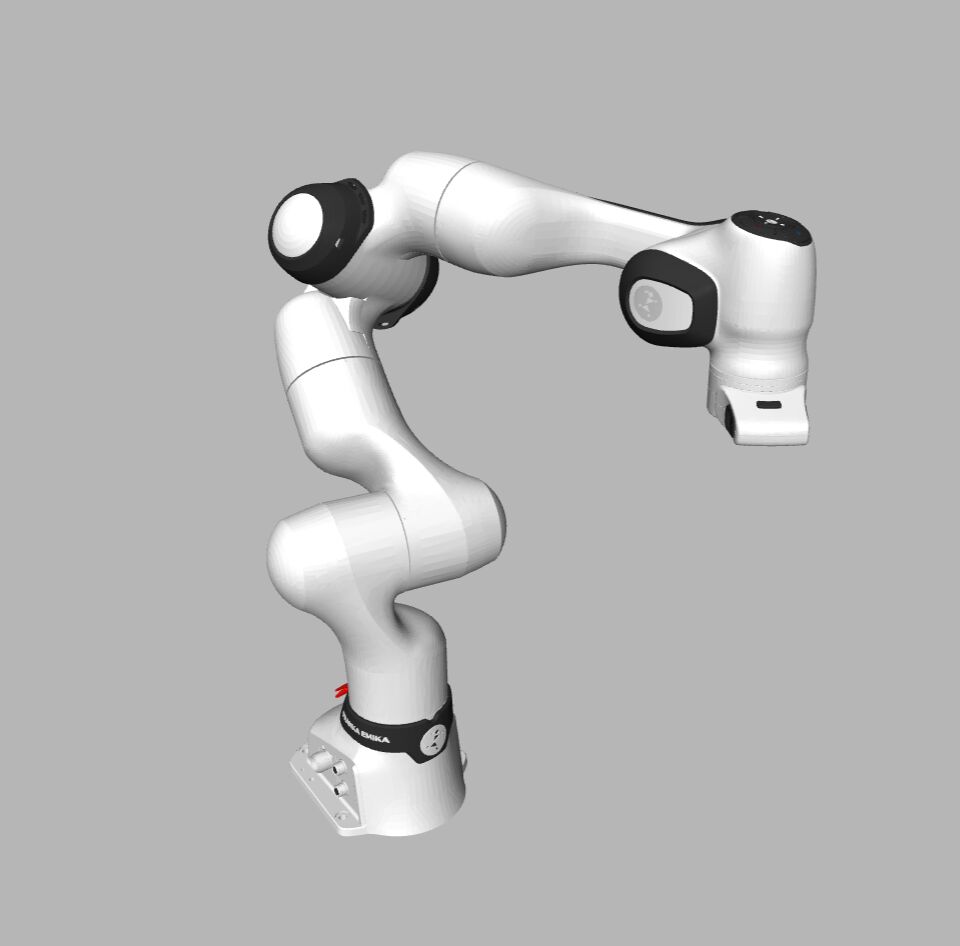}
  \end{minipage}
  \begin{minipage}[b]{0.24\linewidth}
    \centering
    \includegraphics[width=\linewidth]{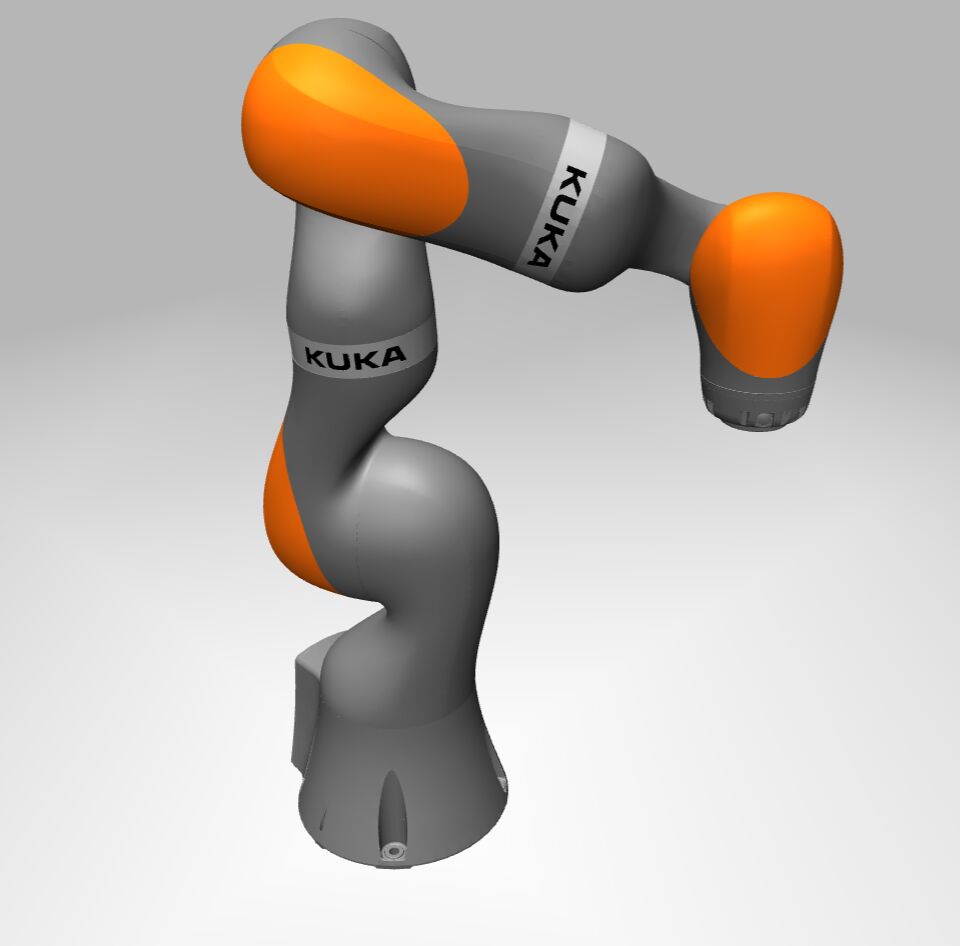}
  \end{minipage}
  \begin{minipage}[b]{0.24\linewidth}
    \centering
    \includegraphics[width=\linewidth]{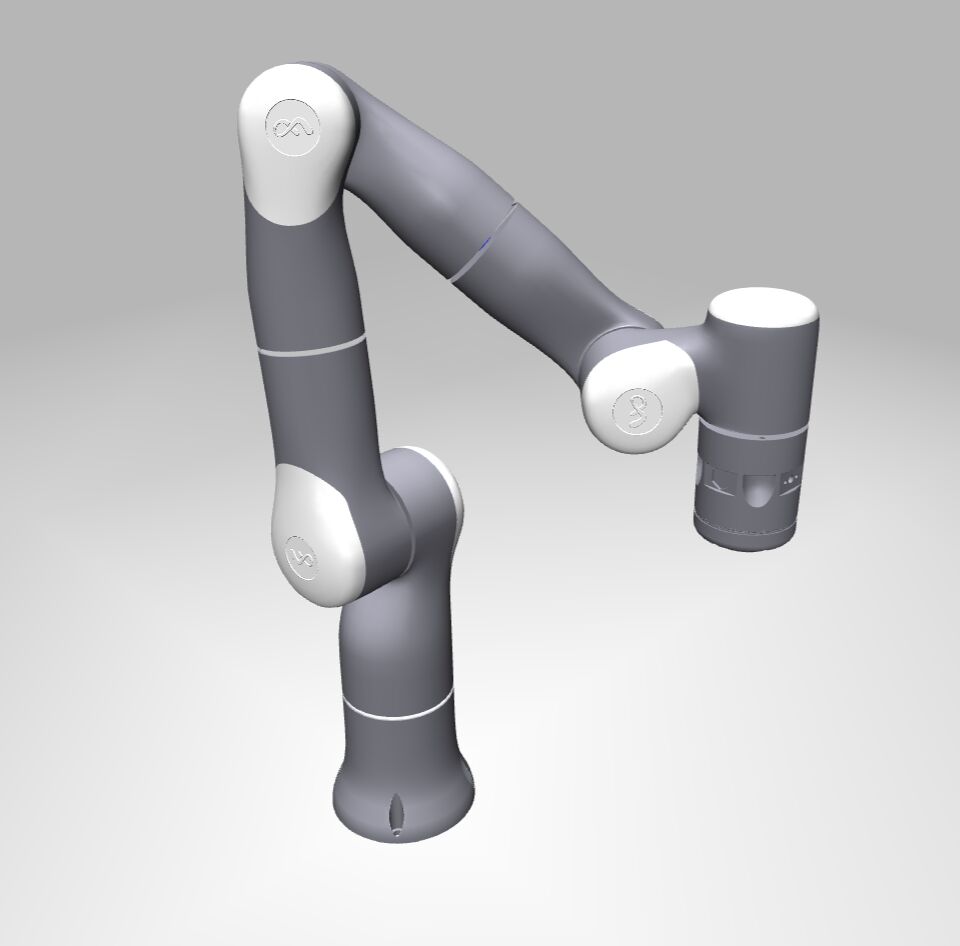}
  \end{minipage}
  \begin{minipage}[b]{0.24\linewidth}
    \centering
    \includegraphics[width=\linewidth]{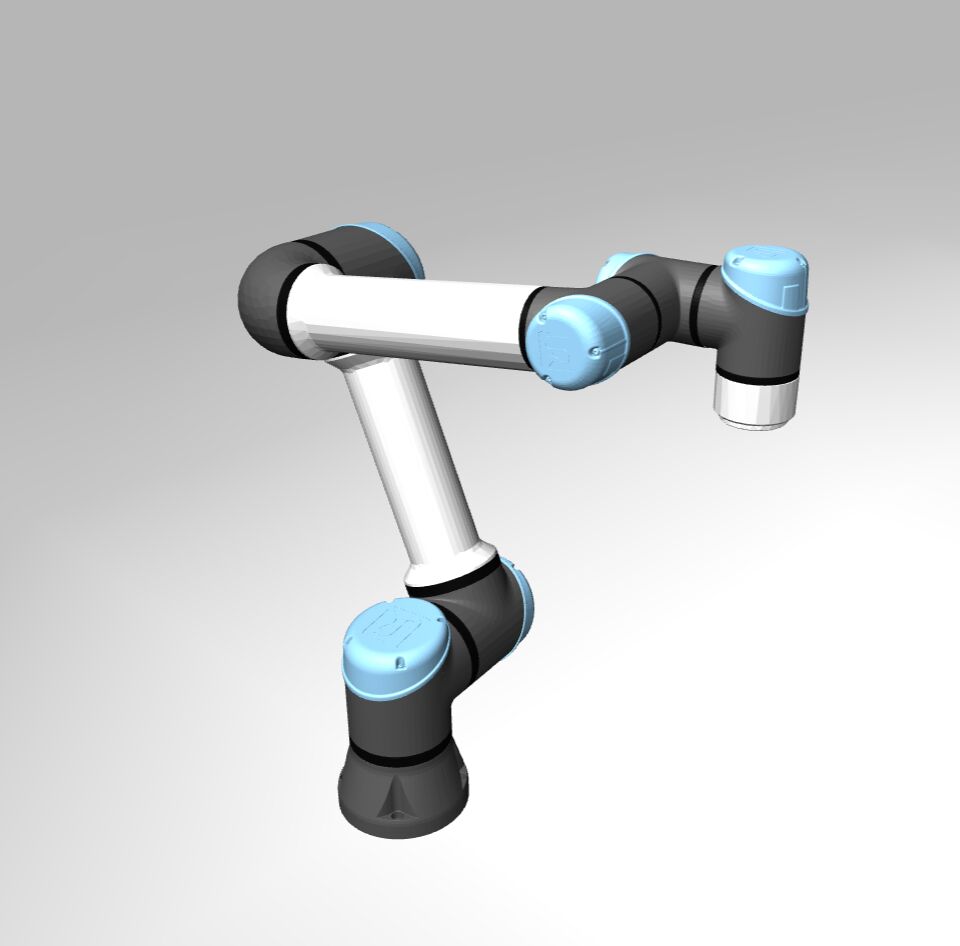}
  \end{minipage}
  \caption{Cross-platform deployment of the same Cartesian impedance plugin on simulated FR3, KUKA LBR iiwa14, Flexiv Rizon 4s, and UR5e manipulators, requiring only robot-specific configuration changes.}
  \label{fig:robot_showcase}
\end{figure}

The same Cartesian impedance plugin and command interface were deployed on a physical FR3 and simulated FR3, KUKA LBR iiwa14, Flexiv Rizon 4s, and UR5e manipulators (Fig.~\ref{fig:robot_showcase}). Each robot was commanded to reach a common task-space target. The controller implementation and command schema remained unchanged, requiring only robot-specific URDF descriptions, joint and frame definitions, and deployment configurations. This demonstrates portability across six- and seven-degree-of-freedom manipulators.

\subsection{Curved Surface Following}

Task-aligned compliance was evaluated through surface following along a curved fixture, representative of contact-rich processes such as grinding, polishing, and deburring. Such tasks require compliant contact along the surface normal while maintaining stiffness in the tangential directions.


Following \cite{Nemec2018ComplianceFS}, the external task source constructs a moving compliance frame, e.g., the Frenet--Serret frame \cite{Ravani2006}, with
\begin{equation}
\mathbf{R}_{\mathrm{task}} =
\begin{bmatrix}\mathbf{t} & \mathbf{n} & \mathbf{b}\end{bmatrix},
\,
\mathbf{K}_{\mathrm{pos}} =
\mathbf{R}_{\mathrm{task}}\mathbf{K}_{\mathrm{pos,task}}
\mathbf{R}_{\mathrm{task}}^\top,
\end{equation}
where $\mathbf{t}$, $\mathbf{n}$, and $\mathbf{b}$ are the tangent, normal, and binormal unit vectors, respectively. Damping is transformed analogously. The resulting matrices are supplied through the proposed command interface alongside the motion references.

The experiment used a 3D-printed, approximately D-shaped fixture with a maximum lateral deviation of 30~mm over a 180~mm path. The plotted trials used translational stiffness values of $(1200,50,1200)$~N/m. An inward force of 5N was applied in the normal direction. Fixed gains applied these values along the robot $x$, $y$, and $z$ axes, whereas task-aligned gains applied them along the estimated tangent, normal, and binormal. 

\begin{figure}[t]
\centering
\begin{minipage}[t]{0.24\linewidth}
\centering
\includegraphics[width=\linewidth]{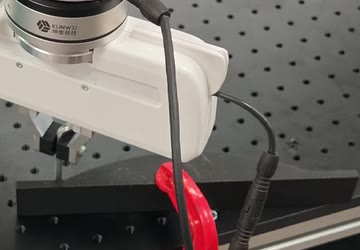}
\end{minipage}%
\hfill
\begin{minipage}[t]{0.24\linewidth}
\centering
\includegraphics[width=\linewidth]{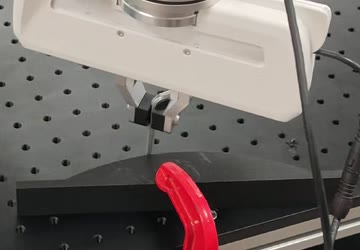}
\end{minipage}%
\hfill
\begin{minipage}[t]{0.24\linewidth}
\centering
\includegraphics[width=\linewidth]{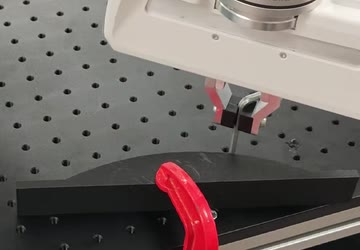}
\end{minipage}%
\hfill
\begin{minipage}[t]{0.24\linewidth}
\centering
\includegraphics[width=\linewidth]{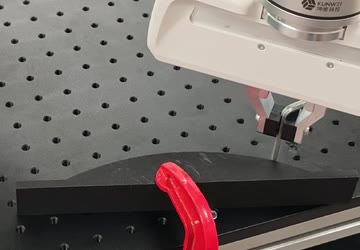}
\end{minipage}\par\vspace{1mm}
\begin{minipage}[t]{0.24\linewidth}
\centering
\includegraphics[width=\linewidth]{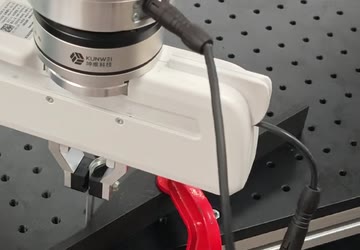}
\end{minipage}%
\hfill
\begin{minipage}[t]{0.24\linewidth}
\centering
\includegraphics[width=\linewidth]{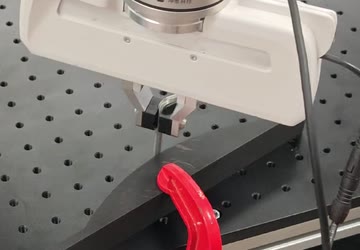}
\end{minipage}%
\hfill
\begin{minipage}[t]{0.24\linewidth}
\centering
\includegraphics[width=\linewidth]{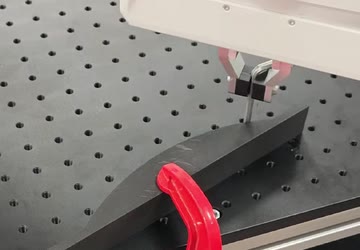}
\end{minipage}%
\hfill
\begin{minipage}[t]{0.24\linewidth}
\centering
\includegraphics[width=\linewidth]{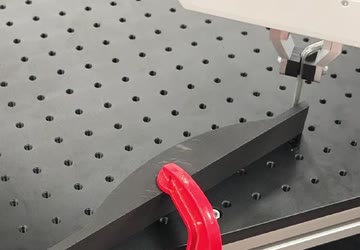}
\end{minipage}
\caption{Snapshots from two recorded surface-following traversals, shown from left to right. The rows show different fixture orientations.}\label{fig:fixture_sequence}
\end{figure}

\begin{figure}[t]
\centering
\includegraphics[width=\linewidth]{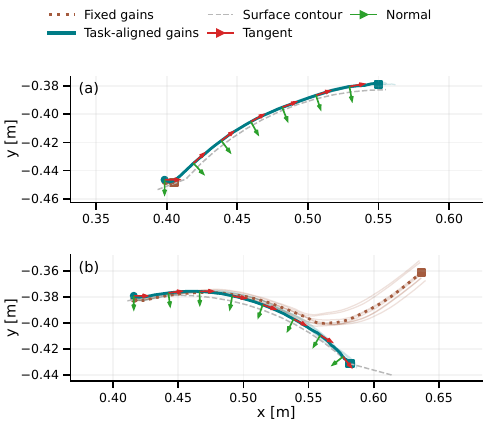}
\caption{Surface following with fixed gains (dotted) and task-aligned gains (solid) in two experimental configurations. Bold traces are representative path medoids; faint traces show the other retained runs. Red and green arrows indicate the estimated tangent and normal. Gray line shows fixture conturs (unknown to the controller).}\label{fig:robot_xy_plot}
\end{figure}

The comparison was repeated in two experimental configurations (Fig.~\ref{fig:fixture_sequence}). Each panel summarizes eight trials with fixed gains and eight with task-aligned gains.
Task-aligned gains produced repeatable curved trajectories,
whereas fixed gains resulted in early stalling in the first
configuration and diversion from the surface-following path
in the second (Fig.~\ref{fig:robot_xy_plot}, below). These observations demonstrate the influence
of compliance-frame orientation on task execution while
preserving the stiffness eigenvalues. The supplementary
video illustrates the contrasting traversal outcomes.
\section{Conclusions}

This paper presents a modular, robot-agnostic framework for compliant control in ROS 2. A wrapper--plugin architecture separates control-law implementations from ROS integration and robot-specific infrastructure, enabling researchers to develop and deploy controllers without implementing the surrounding execution layer, including through block-diagram-based development tools. Standardized joint and Cartesian command interfaces allow high-level applications to update motion references and compliance parameters online, including the orientation of principal compliance directions in task-dependent frames that follow local task geometry.
The same Cartesian impedance implementation was simulated across manipulators with distinct kinematic and dynamic configurations, while real-robot experiments demonstrated the benefits of task-aligned compliance in contact-rich manipulation. 
Beyond the showcased experiments, the framework targets applications in which motion and compliance must be adapted concurrently by high-level command sources, such as teleoperation systems and learned policies that generate motion and interaction parameters during execution.


\section*{Appendix}
\label{sec:supplement}
The framework is released as open-source under Apache-2.0 licence. Table~\ref{tab:supp_arch} identifies the resources and explains how readers can use them to set up the controller, explore its interfaces, and reproduce the deployment workflow. 

\begin{table*}
\caption{Description of Supplementary Materials}
\label{tab:supp_arch}
\begin{tabular*}{\textwidth}{@{\extracolsep{\fill}}p{0.25\textwidth}p{0.7\textwidth}}
Resource & Description\\
\hline
\href{https://github.com/smihael/compliant_controllers}{compliant\_controllers (repository)} & Package implementing joint and Cartesian controller wrappers with interchangeable ROS-independent impedance-control plugins. Provides real-time command buffering, controller lifecycle management, gravity/friction/payload compensation, soft joint-limit handling, torque validation, filtering, and rate limiting. Control laws can be exchanged without modifying the wrappers or hardware integration.\\
\hline
\href{https://github.com/smihael/ros2_control_robot_dynamics/}{ros2\_control\_robot\_dynamics (repository)} & Package providing a robot-agnostic URDF/Pinocchio model library for forward kinematics, end-effector poses, Jacobians, and joint-space dynamics. Supports configurable joint and frame definitions, enabling controller implementations to be reused across robot models.\\
\hline
\href{https://github.com/smihael/compliant_controllers_msgs/}{compliant\_controllers\_msgs (repository)} & Interface package defining standardized joint and Cartesian command messages. Cartesian commands include pose/twist references, translational/rotational stiffness and damping, feedforward wrench, nullspace targets/gains, and compliance-frame orientation. Joint commands include position/velocity references, stiffness/damping, and feedforward torque.\\
\hline
\href{https://github.com/smihael/compliant_controllers_simulink_pipeline}{compliant\_controllers\_simulink\_pipeline (repository)} & MATLAB/Simulink-based controller development pipeline providing model templates, structured command/state buses, and Embedded Coder scripts. Generates dynamically loadable control-law plugins compatible with the controller wrappers, along with validation utilities.\\
\hline
\href{https://github.com/smihael/compliant_controllers_demos}{compliant\_controllers\_demos\allowbreak (repository)} & Demonstration packages providing executable launch files, controller configurations, and Docker/Compose environments for physical FR3 deployment with various peripheral configurations, as well as FR3, iiwa14, UR5e, and Rizon 4s simulations in Gazebo and MuJoCo. Includes usage examples and step-by-step tutorials.\\
\hline
Demos: examples/surface\_following & Curved-fixture simulation with a Jupyter notebook and command-generation script for online task-aligned compliance, enabling replication of the physical experiments in a simulated environment.\\
\hline
supplementary-video.mp4 & Physical surface-following demonstrations comparing fixed and task-aligned compliance gains, illustrating successful traversal and stalled or diverted motion.\\
\hline
\end{tabular*}
\end{table*}


\bibliography{references}

\end{document}